\documentclass[10pt,twocolumn,letterpaper]{article}

\usepackage{iccv}
\usepackage{times}
\usepackage{epsfig}
\usepackage{graphicx}
\usepackage{amsmath}
\usepackage{amssymb}
\usepackage{multirow}
\usepackage[table]{xcolor}
\usepackage{subfig}
\usepackage{booktabs}
\usepackage{color}
\usepackage{authblk}
\makeatletter
\renewcommand\AB@affilsepx{ \qquad  \protect\Affilfont}
\makeatother

\definecolor{green}{RGB}{0,128,0}
\definecolor{brown}{RGB}{128,0,41}

\usepackage[breaklinks=true,bookmarks=false]{hyperref}

\iccvfinalcopy 

\def\iccvPaperID{830} 

\begin{document}
\pagestyle{empty}
\title{FoveaNet: Perspective-aware Urban Scene Parsing}

\vspace{-2cm}
\author{Xin Li$^{1,2}$ \ Zequn Jie$^{3}$ \ Wei Wang$^{4,2}$ \ Changsong Liu$^{1}$ \ Jimei Yang$^{5}$ \\ Xiaohui Shen$^{5}$ \ Zhe Lin$^{5}$ \ Qiang Chen$^{6}$ \ Shuicheng Yan$^{2,6}$ \ Jiashi Feng$^{2}$ \\
\small $^1$ Department of EE, Tsinghua University, $^2$ Department of ECE, National University of Singapore\\
\small $^3$ Tencent AI Lab \ $^4$ University of Trento \ $^5$ Adobe Research \ $^6$ 360 AI institute \\
\small \{lixincn2015,zequn.nus\}@gmail.com  \ wei.wang@unitn.it \ lcs@ocrserv.ee.tsinghua.edu.cn \\ \{jimyang,xshen,zlin\}@adobe.com \ \{chenqiang-iri,yanshuicheng\}360.cn \ elefjia@nus.edu.cn
}

\maketitle
\thispagestyle{empty}


\begin{abstract}
	\vspace{-0.4cm}
Parsing urban scene images benefits many applications, especially self-driving. Most of the current solutions employ generic image parsing models that treat all scales and locations in the images equally and do not consider the geometry property of car-captured urban scene images. Thus, they  suffer from heterogeneous object scales caused by perspective projection of cameras on actual scenes and inevitably encounter parsing failures on distant objects as well as other 
boundary  and recognition errors.
In this work, we propose a  new FoveaNet model  to fully exploit the  perspective geometry of scene images  and address the common failures of generic parsing models. 
FoveaNet estimates the perspective geometry of a scene image through a convolutional network which integrates supportive evidence from contextual objects within the image.	 
Based on the perspective geometry information, FoveaNet ``undoes'' the camera perspective projection~\textemdash~analyzing regions in the space of the actual scene, and thus provides  much more reliable parsing results. Furthermore, to effectively address the recognition errors, FoveaNet introduces a new dense CRFs model that takes the perspective geometry as a prior potential. We evaluate FoveaNet on two urban scene parsing datasets, Cityspaces and CamVid, which demonstrates that FoveaNet can outperform all the well-established baselines and provide new state-of-the-art performance.
	\vspace{-0.6cm}
\end{abstract}

\section{Introduction}
	\vspace{-0.2cm}

Urban scene parsing is a heated research topic that finds application in many fields, especially self-driving. It aims to predict the semantic category for each pixel within a scene image captured by car mounted cameras, which enables self-driving cars to perform reasoning about both the overall scene background and the individual objects moving in front of the cars. 
\begin{figure}[t]
	\vspace{-0.3cm}
	\centering
	\includegraphics[width=0.8\linewidth, height=0.65 \linewidth]{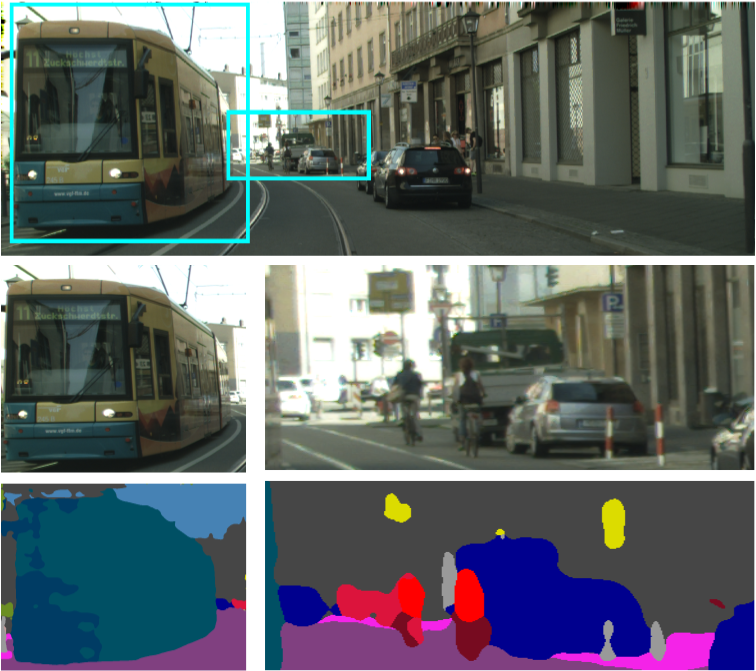}
	\caption{\small Illustration of our motivation. Top two rows: a scene image with perspective geometry and its two zoomed-in regions. Bottom two rows: typical failures in urban scene parsing. Left: ``broken-down'' error on objects of large scales (the bus). Right: boundary errors on objects of small scales.}
	\label{fig:perspective_demo}
	\vspace{-7mm}
\end{figure}

Recent progress in urban scene parsing is mostly driven by the advance of deep learning. 
Deep convolutional neural network (CNN) based parsing algorithms~\cite{long2015fully,liang2015semantic} have demonstrated remarkable performance on several semantic parsing benchmarks~\cite{deng2009imagenet,cordts2016cityscapes,lin2014microsoft}.
However, directly applying the generic CNN based image parsing models usually leads to unsatisfactory results on urban scene images for self-driving cars, since they ignore the important perspective geometry of scene images.

As captured by ego-centric cameras, perspective projection from actual scenes to the image plane changes the object scales: a nearby car seems much bigger than a car far away, even though they have the same scale in reality. The top row in Figure~\ref{fig:perspective_demo} illustrates such a perspective geometry structure within a scene image. Generic parsing models do not take such heterogeneous object scales into consideration. Consequently, they do not perform well on parsing distant objects (of small scales), and boundary and recognition errors are introduced. See the parsing result marked with the small box in Figure~\ref{fig:perspective_demo}. In addition, objects that are near to the camera and usually distributed within the peripheral region have relatively large scales. Generic parsing models tend to ``break down'' a large-scale object into several pieces of similar classes, as shown in the parsing result marked with the big box in Figure~\ref{fig:perspective_demo}. Both of the above problems  are from  ignoring the perspective geometry.

Therefore, we propose a novel FoveaNet  to handle  heterogeneous scales in urban scene parsing by considering the perspective geometry. FoveaNet works like the fovea of human eyes: the center of the vision field (fovea region)  is focused on and the visual acuity is the highest. 
Through localizing ``the fovea region" during parsing, FoveaNet ``undoes'' the camera perspective projection by scale normalization and parses regions at suitable scales.

Specifically, FoveaNet employs a perspective estimation network to infer the overall perspective geometry and output dense perspective scores for each individual pixel, indicating the nearness of a pixel to the vanishing point. Objects with large perspective scores are usually small in the projected scene image. To address the unsatisfactory performance on parsing distant objects, FoveaNet performs \emph{scale normalization} on the fovea region that consists of small-scale objects. 
Then the parsings of small distant objects and large near objects are untangled by a perspective-aware parsing scene network, and boundary errors induced by small scale objects are reduced.

To address the ``broken-down'' issues with parsing large objects, 
FoveaNet employs a new perspective-aware dense CRFs model that takes as input the perspective information and outputs different potentials on the pixels of different perspective scores. The proposed CRFs smooths the pixels from distant objects with large perspective scores more slightly than on the large objects. Through this adaptive strategy, the proposed CRFs are able to handle the ``broken-down'' errors and meanwhile avoid over-smoothing on small objects. We evaluate the proposed FoveaNet on two challenging datasets, Cityspaces and CamVid, and prove that it can provide new state-of-the-art performance on urban scene parsing problems.
We make following contributions to urban scene parsing:
\begin{itemize}
	\vspace{-2mm}
	\setlength\itemsep{0em}
	\item We propose to consider perspective geometry in urban scene parsing and introduce a perspective estimation network for learning the global perspective geometry of urban scene images.
	\vspace{-1mm}
	\item We develop a perspective-aware parsing network that addresses the scale heterogeneity issues well for urban scene images and  gives
	\vspace{-1mm} accurate parsing on small objects crowding around the vanishing point.
	\item We present a new perspective-aware CRFs model that is able to reduce the typical ``broken-down'' errors in parsing peripheral regions of a scene image. 
\end{itemize}

\vspace{-3mm}
\section{Related Work}
\vspace{-2mm}
\label{sec:related_work}

\paragraph{Semantic Parsing}
Recently, deep learning has greatly stimulated the progress on parsing tasks.
Among CNN based algorithms, the Fully Convolutional Network (FCN)~\cite{long2015fully} and the DeepLab model~\cite{liang2015semantic} have achieved most remarkable success. Afterwards, various approaches have been proposed to combine the strengths of FCN and CRFs~\cite{zheng2015conditional,lin2015efficient}, or to refine predictions by exploiting  feature maps output by more bottom layers~\cite{pinheiro2016learning,ghiasi2016laplacian}. 
A common way to deal with scale issues in parsing is to zoom in the input images~\cite{farabet2013learning,mostajabi2015feedforward,dai2015boxsup,lin2016exploring,CY2016Attention}. The input images are rescaled to multiple scales and processed by a shared deep network~\cite{dai2015boxsup,lin2016exploring,CY2016Attention}. 
More recently, Xia~\emph{et al.}~\cite{xia2016zoom} addressed the scale issues in the scenario of object parsing by  ``zoom and refine''.
However, it is not suitable for urban scene parsing.
Our FoveaNet differs from end-to-end trained attention models which learn black-box localization functions~\cite{sharma2015action,xu2015show,mnih2014recurrent,juefei2016deepgender}. Instead, FoveaNet explicitly models the visible geometry structure for fovea region localization and better fits the urban scene parsing task.

\begin{figure*}[t]
	\vspace{-3mm}
	\centering
	\includegraphics[width=0.88\linewidth]{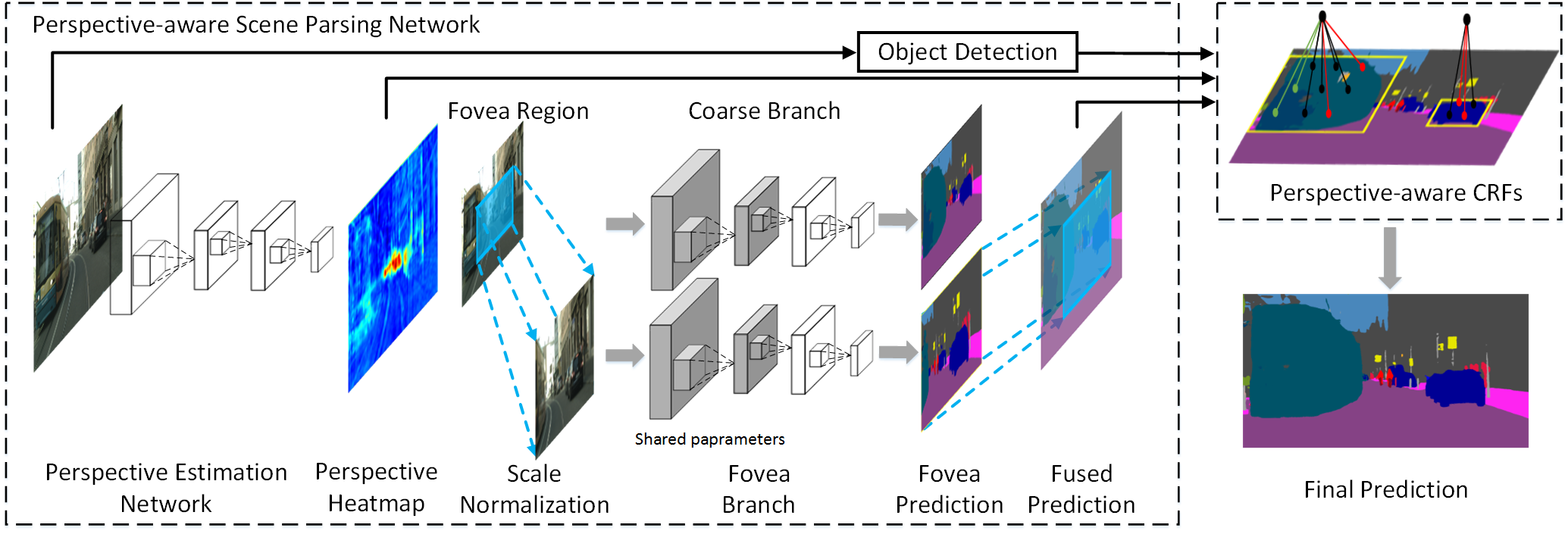}
	\vspace{-2mm}
	\caption{\small Architecture overview of FoveaNet. FoveaNet consists of a perspective-aware parsing network and perspective-aware CRFs. With the perspective estimation network (PEN), FoveaNet infers the global perspective geometry by producing a heatmap. Based on the perspective heatmap,  FoveaNet localizes a fovea region (cyan rectangle) where small distant objects crowd. FoveaNet performs scale normalization on the fovea region, on which it produces a finer parsing via the Fovea branch. This result is then fused with the parsing from a coarse branch  into the final prediction. The perspective-aware CRFs take input the fused parsing result, the perspective heatmp as well as object detection results, and output the final parsing result. {Best viewed in color.}}
	\label{fig:pipeline}
	\vspace{-5mm}
\end{figure*}


\vspace{-5mm}

\paragraph{Perspective Geometry in Urban Scenes} As 3D perspective geometry is a key property of urban scene images, several works consider modeling 3D geometric information as an additional feature for scene understanding~\cite{sturgess2009combining,ladicky2014pulling,hoiem2005geometric,hoiem2008putting,zhang2015monocular}. 
Sturgess~\emph{et al.}~\cite{sturgess2009combining} made use of geometric features in road scene parsing, which are computed using 3D point clouds.
Hoiem~\emph{et al.}~\cite{hoiem2005geometric} modeled geometric context through classifying pixels into different orientation labels.
Some others infer proper object scales with perspective geometry~\cite{hoiem2008putting,zhang2015monocular,ladicky2014pulling}. For example, Hoiem~\emph{et al.}~\cite{hoiem2008putting} established the relationship between camera viewpoint and object scales, and used it as a prior for an object proposal.
Ladicky~\emph{et al.}~\cite{ladicky2014pulling} trained a classifier with hand-crafted features to jointly solve semantic parsing and depth estimation. Training samples are transformed into the canonical depth, due to the observation that performance is limited by the scale misalignment due to the perspective geometry. 
All of the methods above are based on hand-crafted features rather than deep learning. 

\vspace{-3mm}
\section{The Proposed FoveaNet}
\label{sec:foveanet}

\subsection{Overview}
The basic idea of FoveaNet is estimating the perspective geometry of an urban scene image and parsing regions at suitable scales, instead of processing the whole image at a single scale.
The overall architecture of FoveaNet is illustrated in Figure~\ref{fig:pipeline}. The FoveaNet consists of two  components, \emph{i.e.}, the perspective-aware parsing net  and the perspective-aware CRFs.

The perspective-aware parsing  net aims at better parsing small scale objects crowding around the vanishing point by exploiting the image inherent perspective geometry. We propose a perspective estimation network (PEN) to estimate the perspective geometry by predicting a dense perspective heatmap, where a pixel of an object nearer to the vanishing point would have a larger value. Thus PEN provides clues to locate a \emph{fovea region} within which most small scale objects crowd. The fovea region is then re-scaled and receives finer processing by the parsing net, \emph{i.e.} a two-branch FCN.
In this way, small distant objects are untangled from large near objects for parsing.

The perspective-aware CRFs aim at addressing ``broken-down'' errors when parsing the peripheral region of a scene image.
Within this new CRFs model, we introduce a spatial support compatibility function that incorporates the perspective information from PEN, and facilitates the parsing by imposing adaptive potentials at different locations with different perspective heatmap scores. Only the regions confidently from the same object are processed by the CRFs. Small distant objects will be smoothed in a lighter way than the large near objects. The ``broken-down'' errors in peripheral regions can be alleviated effectively.
We now proceed to introduce each component of FoveaNet, respectively.

\subsection{FCN in FoveaNet}
\label{subsec:fcn}
FoveaNet is based on the fully convolutional network~(FCN)~\cite{long2015fully} for parsing the images.
As a deeper CNN model benefits more for the parsing performance, we here follow Chen \emph{et al.}~\cite{CP2016Deeplab} and use the vanilla ResNet-101~\cite{he2016deep} to initialize the FCN model in FoveaNet.
We observe that preserving high spatial resolution of feature maps is very important for accurately segmenting small objects within scenes. Therefore, we disable the last down-sampling layer by setting its stride as $1$.
This increases the size of the feature map output by res$5\_c$ to $1 / 16$ of the input image size (without this modification the size of the output feature map is only $1 / 32$ of the input image size).

\begin{figure}[t]
	\vspace{-1mm}
	\centering
	\includegraphics[width=0.82\linewidth]{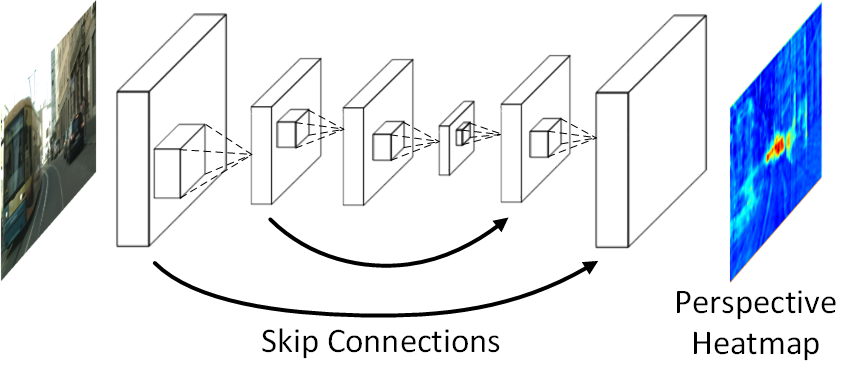}
	\vspace{-2mm}
	\caption{\small Architectural overview of the perspective estimation network (PEN). PEN has a similar network structure as the FCN.
	Given an input scene image, PEN produces a one channel heatmap indicating (roughly) the nearness to the vanishing point at pixel-level.}
	\label{fig:fcn}
	\vspace{-4mm}
\end{figure}

\subsection{Perspective-aware Scene Parsing Network}
\label{subsec:pen}
FoveaNet localizes the fovea region with proper scales and concentrates on the localized fovea region to normalize the various object scales. To this end, a perspective estimation network  is used to estimate the overall perspective geometry of a scene image and localize the region (roughly) centered at the vanishing point where most of small scale objects crowd.
PEN then works together with a two-branch FCN as a perspective-aware scene parsing network.

\vspace{-4mm}
\paragraph{Training PEN} PEN  has a same structure as the baseline FCN model, as shown in Figure~\ref{fig:fcn}.
Our ground truth takes the form of a heatmap: a larger value in the heatmap indicates a higher possibility of small objects to crowd. As it is not easy to estimate the vanishing point of a scene image correctly (sometimes the vanishing point may be invisible or not exist in the image), we use the object scale as a clue to roughly estimate the position of the vanishing point and the perspective geometry. 

For training PEN, we formulate the ground truth heatmap of an image as follows:
\begin{align}
\vspace{-2mm}
H_{i}^{(n)} &= \dfrac{AveSize(\ell(m))}{Size(m)}, \text{where } i \in \text{instance } m, \nonumber \\
G_{i} &= \frac{1}{N}\sum_{n=1}^{N} H^{(n)}_{i}, \quad V_{i}^{(n)} =  H_{i}^{(n)}  + \delta \times G_{i},
\label{eqn:gt_heatmap}
\vspace{-4mm}
\end{align}
In the above equation, $m$ denotes an object instance in the $n$-th image, and $i$ indexes a pixel from this instance. $\ell{m}$ denotes the category label of instance $m$. ${AveSize}(\ell(m))$ denotes the category-level average instance size. Thus $H_{i}^{(n)}$, \emph{i.e.} the value of pixel $i$ in the $n$-th heatmap $H^{(n)}$, depends on the ratio of the category-level average instance size over the current instance size $Size(m)$. Global perspective score prior $G_i$ for the $i$-th pixel is the average value over all the $N$ heatmaps. The ground truth $V_{i}^{(n)}$ for training PEN is formulated by weighted summing both the image specific characteristics $H_{i}^{(n)}$ and the global average $G_{i}$, being traded-off by a parameter $\delta$.

\begin{figure}[!tb]
	\centering
	\includegraphics[width=80mm,height=50mm]{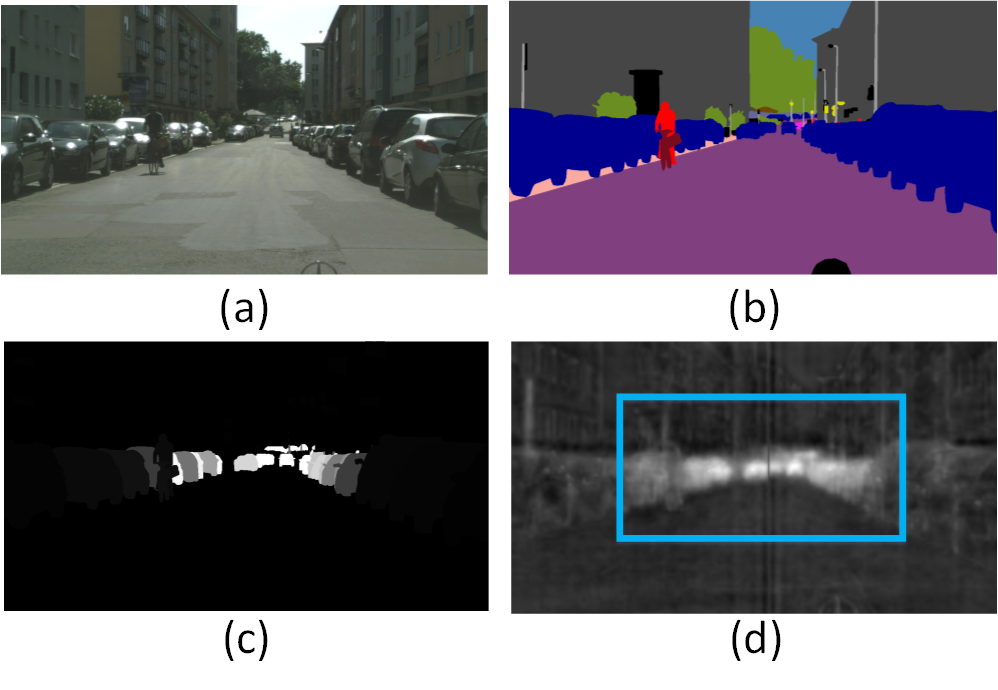}
	\vspace{-4mm}
	\caption{\small Illustration of perspective heatmap estimation.  (a) An urban scene image. (b) Parsing ground truth. (c) Ground truth perspective heatmap generated by Eqn.~\eqref{eqn:gt_heatmap}. (d) Estimated perspective heatmap and detected Fovea region from PEN. Fovea region with maximal response is highlighted in cyan. Best viewed in color.}
	\label{fig:perspective_heatmap}
	\vspace{-4mm}
\end{figure}

PEN is trained by minimizing a smoothed $\ell_1$ loss \cite{girshick2015fast} between the produced heatmap based on raw images and the  ground truth heatmap. 
Figure~\ref{fig:perspective_heatmap} illustrates the result of PEN. Figure~\ref{fig:perspective_heatmap} (a) shows a training urban scene image with perspective geometry and Figure~\ref{fig:perspective_heatmap} (b) shows its ground truth parsing map.
We follow Eqn.~\eqref{eqn:gt_heatmap} to obtain the ground truth perspective confidence map shown in  Figure~\ref{fig:perspective_heatmap} (c).
From the perspective map estimated by PEN (Figure~\ref{fig:perspective_heatmap} (d)), one can observe that PEN  successfully predicts the overall geometry of the input image~\textemdash~it outputs  larger values for the pixels closer to the vanishing point. 

With this perspective heatmap, FoveaNet localizes the fovea region  with maximal response (highlighted with cyan rectangle).
In our experiments, we  define the size of the fovea region as $1/2$ of the heatmap size.
To locate the fovea region based on the heatmap, FoveaNet passes the heatmap from PEN through an average pooling layer. The receptive field of the maximal pooling result on the heatmap is selected as the fovea region, as illustrated by the cyan boxes in Figure~\ref{fig:pipeline} and Figure~\ref{fig:perspective_heatmap} (d).

\vspace{-3mm}
\paragraph{Discussion}
 Another choice for estimating perspective information is to estimate depth information from a single image \cite{garg2016unsupervised,eigen2014depth}.
 However, the single image depth prediction results are not discriminative for localizing distant objects. In contrast, PEN can produce a heatmap with distinguishable per-pixel scores, leading to more precise  fovea region localization.
Therefore, we use the method introduced above to estimate the perspective geometry and train PEN.  A qualitative comparison between predicted depth \cite{garg2016unsupervised} and our estimated perspective heatmap on Cityscapes dataset is provided in  supplementary material.

\vspace{-2mm}
\paragraph{Perspective-aware Scene Parsing} FoveaNet performs \emph{scale normalization} to achieve better parsing performance on objects of heterogeneous scales.
After localizing the fovea region, FoveaNet parses the fovea region and the raw image separately through a two-branch FCN, as shown in  Figure~\ref{fig:pipeline}.
The raw input image passes through the \emph{coarse branch} to produce an overall parsing result. Meanwhile, the fovea region is re-scaled to the original input size and passes through the \emph{fovea branch} to produce finer parsing for the fovea region. 
The two branches have the same structure as the  baseline FCN model and share parameters from conv$1$ to res$3\_3b3$. More architectural details are given in Section~\ref{subsec:fcn}. 
The two-branch FCN is end-to-end trainable by minimizing per-pixel cross-entropy loss.

\subsection{Perspective-aware CRFs}
The perspective-aware scene parsing network can parse the distant objects better by estimating perspective information.
However, another common issue in parsing scene images is that
large objects in peripheral regions of a scene image usually suffer from ``broken-down'' errors, \emph{i.e.}, a large object tends to be broken into several small pieces which are  misclassified into different yet similar classes.
This problem is illustrated in the bottom left subfigure of Figure~\ref{fig:perspective_demo}: some parts of the bus are misclassified into the train, harming the parsing performance on the peripheral region.

Intuitively, it would be beneficial for the final performance to refine the prediction with the aid of appearance features. In object segmentation, dense CRFs are usually applied to the prediction scores produced by FCN, and have shown impressive effects on refining prediction. However, directly applying  dense CRFs to urban scene images does not give satisfactory performance due to heterogeneous scales of objects from the fovea region and the peripheral region.
A dense CRFs model performing well on the peripheral region tends to over-smooth the predictions on small objects from the fovea region, which harms the performance on small objects significantly.

Based on the perspective information from PEN, we propose a new perspective-aware  dense CRFs model to alleviate ``broken-down'' errors.
The CRFs model is trained separately, following the DeepLab model~\cite{liang2015semantic}.
Let $\ell$ denote the label vector for all the pixels, $f_i$ denote the learned representation of the pixel $i$, and $p_i$ denote the 2D-coordinate of the pixel $i$ in the image plane. The energy function of the perspective-aware dense CRFs is defined as
\vspace{-0.2cm}
\begin{equation*}
\mathcal{E}(\ell) = \sum_i \psi_u(\ell_i) + \sum_{i,j} \psi_{p,\text{persp}}(\ell_i,\ell_j).
\vspace{-0.3cm}
\end{equation*}
Here $\psi_u$ is the standard unary potential.
The pairwise potential in our proposed CRFs model has a new form:
\vspace{-0.1cm}
\begin{equation*}
\psi_{p,\text{persp}}(\ell_i,\ell_j) = \mu(p_i,p_j)\nu(\ell_i,\ell_j)  \kappa (f_i,f_j),
\vspace{-0.1cm}
\end{equation*}
where the kernel $\kappa(\cdot,\cdot)$ is the contrast-sensitive two-kernel potential proposed by Krahenbuhl~\emph{et al.}~\cite{KrahenbuhlK11}, and $\nu$ is the Potts label compatibility function. Here  $\mu$ is a new spatial support compatibility function introduced for the perspective-aware CRFs that considers auxiliary object detection results and perspective information provided by PEN:
\vspace{-0.2cm}
\begin{equation}
\label{eqn:perspective}
	\mu(p_i,p_j) = 
	d_k \left (  \frac{\sum_{m\epsilon \widehat{V}}v_{m}}{|\widehat{V}|} / \frac{\sum_{n\epsilon B_k}v_{n}}{|B_k|} \right )
\end{equation}
The object bounding boxes are detected by a Faster-RCNN~\cite{ren2015faster} model. Among them, some bounding boxes $B_k,k=1,2...K$ contain the pixels $p_i$,$p_j$. Then the box $B_k$ with the maximum detection score $d_k$ is selected as the target one.
Here $\widehat{V}$ denotes the estimated heatmap, and $m,n$ index pixels from the estimated heatmap $\widehat{V}$ and bounding box $B_k$ respectively.
This $\mu(p_i,p_j) $ incorporates perspective information as follows. It lowers the weights of the pairwise potential at a bounding box with higher heatmap values. Thus for a small object with a high perspective score, the pairwise potential becomes small (imposing lighter spatial smoothing), and the unary potential plays a major role. 
By focusing on each detection proposal with adaptive perspective weights, the proposed CRFs model effectively alleviates the ``broken-down'' problems and meanwhile avoids over-smoothing the details. 


\section{Experiments}
\label{sec:experiment}
\subsection{Experimental Settings}
We implement FoveaNet using the Caffe library~\cite{jia2014caffe} and evaluate its performance on two urban scene parsing datasets: Cityscapes~\cite{cordts2016cityscapes} and Camvid~\cite{BrostowFC:PRL2008}.
For performing ablation studies on FoveaNet, we employ a vanilla FCN architecture with ResNet-101 being its front-end model as the baseline. It takes raw images as inputs and is trained with per-pixel cross-entropy loss. During testing, it produces parsing results at a single scale. We examine how its performance changes by incorporating different components from FoveaNet, in order to understand the contribution of each component.
FoveaNet is initialized by a modified ResNet-101 network pre-trained on ImageNet (see Section~\ref{subsec:fcn} for more details). We fine-tune the initial model on an individual scene parsing dataset.
The initial learning rate is $0.001$, and is decreased by a factor of $0.1$ after every $20$ epochs for twice. The momentum is $0.9$.


\subsection{Results on Cityscapes}
The Cityscapes dataset \cite{cordts2016cityscapes} is a recently released large-scale benchmark for urban scene parsing. Its images are taken by car-carried cameras and are collected in streets of $50$ different cities. It contains in total $5{,}000$ images with high quality pixel-level annotations. These images are split to $2{,}975$ for training, $500$ for validation and $1{,}525$ for testing.  Cityscapes provides  annotations at two semantic granularities \emph{i.e.}, classes and higher-level categories. Annotations can be divided into 30  classes and 8 higher-level categories. For instance, the classes of \emph{car}, \emph{truck}, \emph{bus} and other $3$ classes are grouped into the \emph{vehicle} category. Among them, 19 classes and 7 categories are used for evaluation.  Our FoveaNet is trained on $2{,}975$ training images, and evaluated on the validation set. Then we add $500$ validation images to fine-tune our model and obtain the test performance.

Following the provided evaluation protocol with the dataset~\cite{cordts2016cityscapes}, we report the performance of compared models in terms of four metrics \emph{i.e.} $\text{IoU}_\text{class}$, $\text{IoU}_\text{category}$, $\text{iIoU}_\text{class}$ and $\text{iIoU}_\text{category}$. Compared with the standard $\text{IoU}_\text{class}$ and $\text{IoU}_\text{category}$, the latter two IoU metrics put more emphasis on the performance on small scale instances.
The resolution of images is $2048\times1024$, which brings a challenge to training deep networks with limited GPU memory. Hence, we use a random image crop of $896\times896$ in training.
For building the perspective-aware CRFs model, we train a Faster-RCNN on Cityscapes with $8$ classes whose ground truth bounding boxes can be derived from instance annotations, including \emph{truck}, \emph{bus}, \emph{motorcycle}.

\begin{table*}[t]
	\centering
	\vspace{-0mm}
	\caption{\small Performance comparison among several variants of FoveaNet on the Cityscapes \emph{validation} set. The metric of $iIoU$ is not applicable for categories of \emph{road} to \emph{bicycle}. Best viewed in color. }
	\vspace{-3mm}
	\label{table:cityscapes_val}
	\footnotesize
	\setlength{\tabcolsep}{2.3pt}
	\begin{tabular}{l|c|ccccccccccccccccccc|cc}
		\multicolumn{1}{c|}{}   &\multicolumn{1}{c|}{\footnotesize Metric}         & \multicolumn{1}{c}{\raisebox{-0.1cm}{\rotatebox[origin=l]{90}{\footnotesize road}}} &  \multicolumn{1}{c}{\raisebox{-0.1cm}{\rotatebox[origin=l]{90}{\footnotesize sidewalk}}} & \multicolumn{1}{c}{\raisebox{-0.1cm}{\rotatebox[origin=l]{90}{\footnotesize building}}} & \multicolumn{1}{c}{\raisebox{-0.1cm}{\rotatebox[origin=l]{90}{\footnotesize wall}}} & \multicolumn{1}{c}{\raisebox{-0.1cm}{\rotatebox[origin=l]{90}{\footnotesize fence}}} & \multicolumn{1}{c}{\raisebox{-0.1cm}{\rotatebox[origin=l]{90}{\footnotesize pole}}} & \multicolumn{1}{c}{\raisebox{-0.1cm}{\rotatebox[origin=l]{90}{\footnotesize tr. light}}} & \multicolumn{1}{c}{\raisebox{-0.1cm}{\rotatebox[origin=l]{90}{\footnotesize tr. sign}}} & \multicolumn{1}{c}{\raisebox{-0.1cm}{\rotatebox[origin=l]{90}{\footnotesize veg.}}} & \multicolumn{1}{c}{\raisebox{-0.1cm}{\rotatebox[origin=l]{90}{\footnotesize terrain}}} & \multicolumn{1}{c}{\raisebox{-0.1cm}{\rotatebox[origin=l]{90}{\footnotesize sky}}} & \multicolumn{1}{c}{\raisebox{-0.1cm}{\rotatebox[origin=l]{90}{\footnotesize person}}} & \multicolumn{1}{c}{\raisebox{-0.1cm}{\rotatebox[origin=l]{90}{\footnotesize rider}}} & \multicolumn{1}{c}{\raisebox{-0.1cm}{\rotatebox[origin=l]{90}{\footnotesize car}}} & \multicolumn{1}{c}{\raisebox{-0.1cm}{\rotatebox[origin=l]{90}{\footnotesize truck}}} & \multicolumn{1}{c}{\raisebox{-0.1cm}{\rotatebox[origin=l]{90}{\footnotesize bus}}} & \multicolumn{1}{c}{\raisebox{-0.1cm}{\rotatebox[origin=l]{90}{\footnotesize train}}} & \multicolumn{1}{c}{\raisebox{-0.1cm}{\rotatebox[origin=l]{90}{\footnotesize mcycle}}} & \multicolumn{1}{c|}{\raisebox{-0.1cm}{\rotatebox[origin=l]{90}{\footnotesize bicycle}}} & \multicolumn{1}{c}{\raisebox{-0.1cm}{\rotatebox[origin=l]{90}{\footnotesize Class}}} & \multicolumn{1}{c}{\raisebox{-0.1cm}{\rotatebox[origin=l]{90}{\footnotesize Category}}} \\ \hline 
		\multirow{2}{*}{FCN Baseline} & IoU & 97.7 & 81.9 & 91.0 & 48.5 & 52.9 & 58.2 & 63.1 & 73.5 & 91.4 & \textbf{61.6} & 94.3 & 78.1 & 56.0 & 93.4 & 57.5 & 81.2 & 66.2 & 60.4 & 74.2 & 72.7 & 87.6 \\ 
		& iIoU & \multicolumn{11}{c}{\cellcolor{gray!20} {------------------}} & 60.4 & 37.8 & 84.8 & 36.2 & 58.5 & 43.1 & 36.9 & 56.4 & 51.8 &72.8 \\ \hline 
		\multirow{2}{*}{+ fixed fovea region} & IoU & 97.8 & 82.8 & 91.3 & 48.0 & 51.0 & 60.8 & 66.7 & 75.6 & 91.7 & 61.3 & \textbf{94.5} & 79.2 & 57.0 & 93.3 & 55.3 & 79.9 & 65.1 & 62.4 & 75.1 & 73.1 & 88.3 \\ 
		& iIoU & \multicolumn{11}{c}{\cellcolor{gray!20} {------------------}} & 62.8 & 46.2 & 86.5 & 36.3 & 59.6 & 43.2 & 42.2 & 59.6 & 54.6 & 75.4 \\ \hline 
		\multirow{2}{*}{+ PEN fovea region} & IoU & 97.8 & 82.9 & 91.4 & 48.6 & 54.3 &\textcolor{green}{ 62.5} & \textcolor{green}{69.0} & \textcolor{green}{77.3} & 91.9 & 60.7 & 94.4 & 80.6 & 60.3 & 93.6 & 56.8 & 80.2 & 60.4 & 65.8 & 76.2 & 73.9 & 88.8 \\ 
		& iIoU & \multicolumn{11}{c}{\cellcolor{gray!20} {------------------}} & 64.7 & 48.6 & 87.2 & 42.8 & 62.2 & 45.3 & \textbf{46.9} & 61.4 & \textcolor{blue}{57.4} & \textcolor{blue}{76.6} \\ \hline \hline
		\multirow{2}{*}{\begin{tabular}[l]{@{}l@{}}+ PEN fovea region\\ \& normal CRFs\end{tabular}} & IoU & 97.7 & 82.7 & 90.7 & 48.7 & 51.5 & \textcolor{orange}{54.1} & \textcolor{orange}{60.7} & \textcolor{orange}{75.3} & 90.9 & 62.9 & 94.5 & 78.9 & 57.7 & 93.3 & \textcolor{brown}{61.8} &  \textcolor{brown}{83.4} &  \textcolor{brown}{70.8} & 65.2 & 74.7 & 73.5 & 87.2 \\ 
		& iIoU & \multicolumn{11}{c}{\cellcolor{gray!20} {------------------}} & 61.6 & 47.0 & 83.3 & 36.2 & 58.5 & 43.7 & 45.7 & 59.0 &54.4 & 73.6 \\ \hline
		\multirow{2}{*}{\begin{tabular}[l]{@{}l@{}}+ PEN fovea region\\ \& depth-aware CRFs\end{tabular}} & IoU & 97.7 & 82.4 & 91.2 & 47.2 & 53.9 & 61.8 & 67.9 & 76.5 & 91.7 & 60.5 & 94.2 & 79.8 & 58.6 & 93.9 & 60.6 &  84.2 &  69.7 & 64.2 & 75.5 & 74.3 & 88.4 \\ 
		& iIoU & \multicolumn{11}{c}{\cellcolor{gray!20} {------------------}} & 63.2 & 46.7 & 87.0 & 40.4 & 60.6 & 44.2 & 42.8 & 59.8 &55.6 & 75.6 \\ \hline

		\multirow{2}{*}{\begin{tabular}[l]{@{}l@{}}+ PEN fovea region \\ \& persp-aware CRFs\end{tabular}} & IoU & \textbf{97.9} & \textbf{83.0} & \textbf{91.5} & \textbf{47.7} & \textbf{54.5} & \textbf{62.8} & \textbf{69.1} & \textbf{77.5} & \textbf{91.9} & 60.9 & 94.4 & \textbf{80.7} & \textbf{60.4} & \textbf{94.4} & \textcolor{red}{\textbf{72.5}} & \textcolor{red}{\textbf{86.2}} & \textcolor{red}{\textbf{72.7}} & \textbf{66.8} & \textbf{76.4} & \textbf{75.9} & \textbf{88.8} \\ 
		& iIoU & \multicolumn{11}{c}{\cellcolor{gray!20} {------------------}} & \textbf{65.1} & \textbf{48.9} & \textbf{87.8} & \textbf{43.0} & \textbf{62.3} & \textbf{49.2} &  46.8 & \textbf{61.6} & \textbf{58.1} & \textbf{76.8} \\ \hline		
	\end{tabular}
	\vspace{-3mm}
\end{table*}

\begin{figure}
  \centering
   \includegraphics[width=0.7\linewidth]{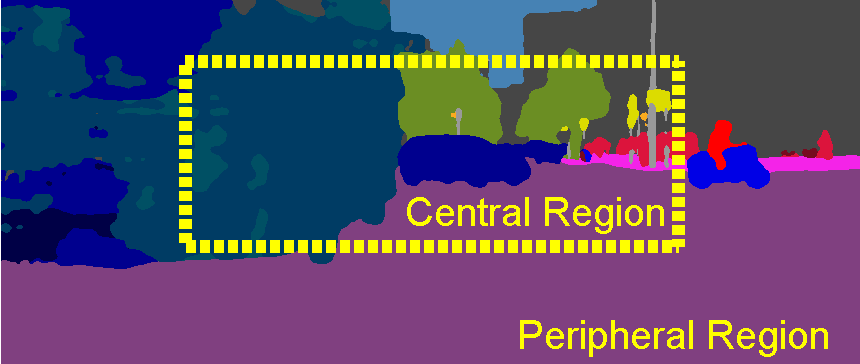}
   \vspace{-2mm}
  \caption{\small Typical parsing result of baseline FCN model. We evaluate FCN on peripheral and central region respectively, to analyze how much a perspective distortion affects urban scene parsing.}
  \label{fig:peripheral_central} 
\vspace{-2mm}
\end{figure}

\vspace{-3mm}
\paragraph{Perspective Distortion} 
We now quantitatively analyze how much perspective distortion affects urban scene parsing and demonstrate perspective distortion is a severe issue for urban scene parsing. 
We evaluate the baseline FCN model (trained on the whole images) on two image sets: one contains only the central region and the other contains only the peripheral region, as illustrated in Figure \ref{fig:peripheral_central}. 
Table \ref{tab:peripheral_central} shows a detailed comparison between these two image sets on \emph{Object} and \emph{Vehicle} category, which consist of $3$ and $4$ classes respectively.
First, we find that the performance on the \emph{Object} category in the central region is much worser than in the peripheral region. More concretely, we find that the $\text{IoU}_\text{category}$ of \emph{Object} drops $10.6\%$ in the central region.
This performance drop comes from the small object scales in the center region caused by perspective distortion. This problem can also be observed from parsing results in Figure~\ref{fig:peripheral_central}. The parsing in the central region lacks enough details.
Second, generic parsing models tend to ``break down'' a large-scale object into several pieces of similar classes, as illustrated in Figure \ref{fig:peripheral_central}. We can observe from Table \ref{tab:peripheral_central} that the $\text{IoU}_\text{Category}$ of \emph{Vehicle} improves $2.7\%$, but corresponding $\text{IoU}_\text{Class}$ deteriorates largely in the peripheral region. This can be largely attributed to misclassification between fine-grained classes, which is reflected by the $\text{IoU}_\text{Class}$ metric. Objects in the peripheral region have an unbalanced larger scale due to perspective distortion. The performance drop on \emph{Vehicle} category is brought by the ``broken-down'' issue.

\begin{table}[]
\centering
\footnotesize
\caption{\small Comparison on \emph{Object} and \emph{Vehicle} category between peripheral and central regions.}
\vspace{-2mm}
\label{tab:peripheral_central}
\begin{tabular}{l|l|cc}
\hline
\multicolumn{2}{c|}{Region}              & peripheral & central \\ \hline \hline
$\text{IoU}_\text{category}$               & object    & 69.7       & 59.1    \\ \hline
\multirow{3}{*}{$\text{IoU}_\text{Class}$} & pole      & 62.8       & 51.1    \\
                            & tr. light & 66.3       & 58.2    \\
                            & tr. sign  & 77.7       & 67.5    \\ \hline\hline
$\text{IoU}_\text{category}$              & vehicle   & 93.3       & 90.6    \\ \hline
\multirow{4}{*}{$\text{IoU}_\text{Class}$} & car       & 94.3       & 91.8    \\
                            & truck     & 48.0       & 66.0    \\
                            & bus       & 78.7       & 83.0    \\
                            & train     & 61.0       & 71.6   \\ \hline
\end{tabular}
\vspace{-4mm}
\end{table}

\begin{figure*}[t]
\vspace{-0mm}
\centering
\includegraphics[width=0.92\linewidth]{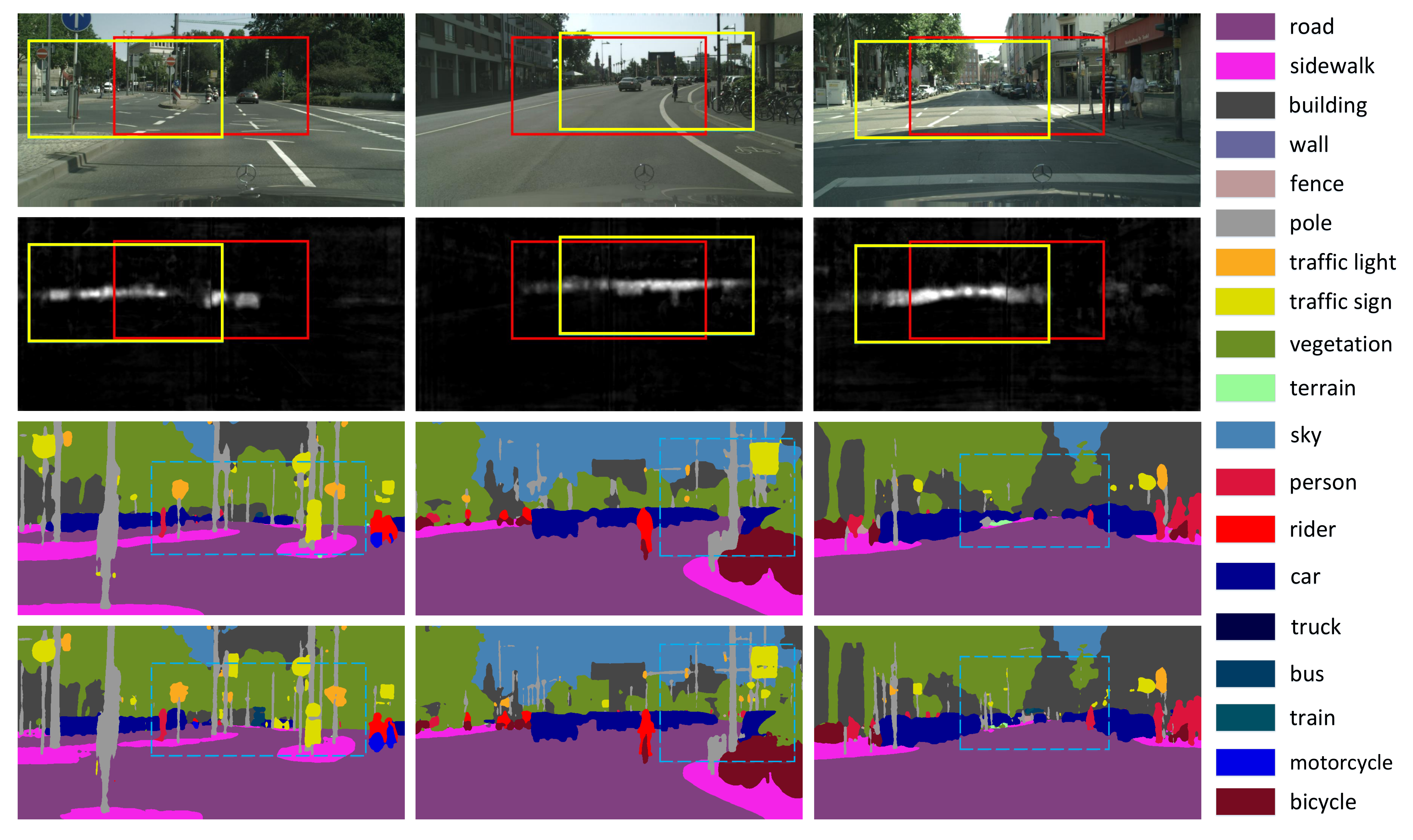}
\vspace{-2mm}
\caption{\small Example parsing results on Cityscapes. 1st-2nd row: urban scene images with two types of fovea regions derived from global prior (red) and PEN (yellow) based on its estimated heatmap (2nd row). 3rd row: parsing result on fovea regions with FCN baseline. 4th row: parsing result on fovea region with FoveaNet. FoveaNet produces more detailed parsing results on small scale objects \emph{e.g.},  \emph{pole}, \emph{traffic light}, \emph{traffic sign}. Best viewed in color.}
\label{fig:qualitative_perspective}
\vspace{-3mm}
\end{figure*}

\vspace{-3mm}
\paragraph{Ablation Analysis}  We now analyze FoveaNet by investigating the effects of each component separately. Table~\ref{table:cityscapes_val} lists the performance of adding each component of FoveaNet to the baseline model (vanilla FCN) on the validation set. We also give a qualitative comparisons in Figure~\ref{fig:qualitative_perspective}. From the results, we can make following observations.

\textit{Perspective-aware Parsing}:
The 2nd row in Figure~\ref{fig:qualitative_perspective} shows that PEN successfully estimates the global perspective geometry. In the heatmap, small scale objects have larger response values (brighter). We compare the fovea region estimated by PEN (yellow rectangle) with a pre-fixed fovea region estimated from the global average (red rectangles; ref.\ Eqn.~\eqref{eqn:gt_heatmap}). Comparing these two fovea regions shows  PEN better localizes the regions covering small objects and is adaptive to different images. For example, the leftmost image presents a road turning left and thus small scale objects 
crowd in the left panel. PEN effectively locates this region but the globally fixed one fails.
  
We also quantitatively compare the benefits of these two fovea region localization strategies in Table~\ref{table:cityscapes_val} (+ fixed fovea region vs.\ + PEN fovea region). One can observe that relying on the fovea regions provided by PEN significantly performs better by a margin of  $2.8\%$ in terms of $\text{iIoU}_\text{class}$. 
Compared with the baseline FCN model, performing perspective-aware parsing with the help of  PEN significantly improves the performance by $5.6\%$ and $3.8\%$ on the instance-level scores $\text{iIoU}_\text{class}$ and $\text{iIoU}_\text{category}$ respectively (highlighted in blue). This verifies  perspective information is indeed beneficial for urban scene parsing.

Figure~\ref{fig:qualitative_perspective} provides more qualitative results.
We visualize the parsing results on the fovea region (from PEN) with FoveaNet or with FCN baseline model (the 4th row and the 3rd row respectively).
One can observe that perspective-aware parsing gives  results with richer details. Particularly, the pole, traffic light and traffic sign are parsed very well. 
This is also confirmed by their $\text{IoU}$ improvement in Table~\ref{table:cityscapes_val} (highlighted in green), which is up to $6\%$. These qualitative and quantitative results clearly validate the effectiveness of the perspective-aware parsing network on  objects of small scales, as it can better address the scale heterogeneity issue in urban scenes.

\textit{Perspective-aware CRFs:}
{Based on the perspective-aware parsing on the fovea region, we further compare perspective-aware CRFs, normal dense CRFs and depth-aware CRFs in Table~\ref{table:cityscapes_val}. The depth-aware CRFs model is  similar to the perspective-aware CRFs model, except that  the perspective heatmap in Eqn.~\eqref{eqn:perspective} is replaced by  single image depth prediction from the method in~\cite{garg2016unsupervised}.
}

\begin{figure*}[t]
\vspace{-2mm}
\centering
\includegraphics[width=0.91\linewidth]{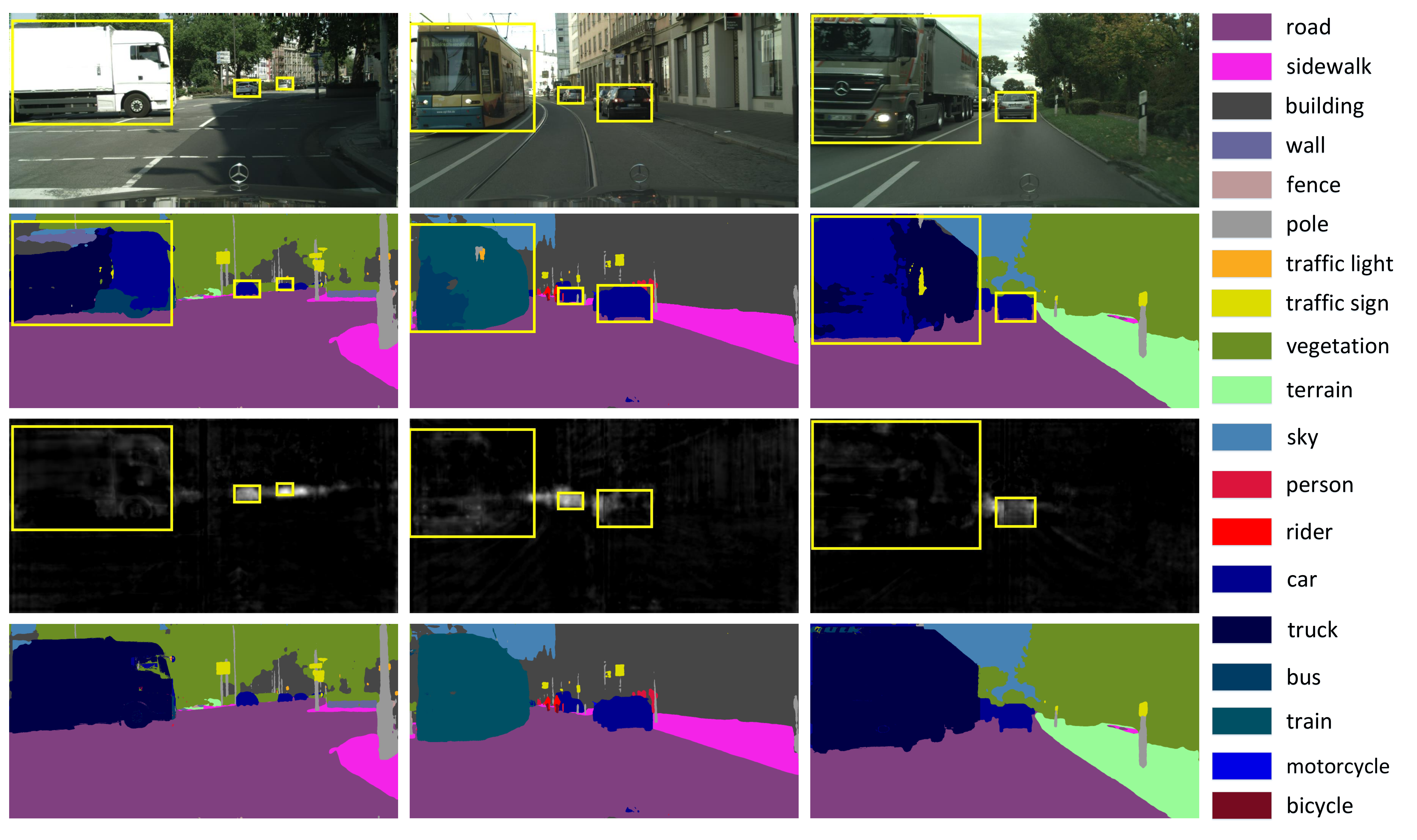}
\vspace{-2mm}
\caption{\small Parsing results of perspective-aware CRFs on Cityscapes validation set. Top: input images with object detection bounding boxes (yellow). 2nd row: parsing results from FCN. Large scale objects in peripheral region present ``broken-down'' errors. 3rd row: the perspective information  by PEN which is integrated into proposed perspective-aware CRFs. Bottom: FoveaNet applies perspective-aware CRFs to remove the ``broken-down'' error. Best viewed in color. 
 }
\label{fig:qualitative_crf}
\vspace{-3mm}
\end{figure*}

We observe that \emph{truck}, \emph{bus} and \emph{train} are the three classes with most severe ``broken-down'' errors. Applying the normal dense CRFs improves the $\text{IoU}_\text{class}$ of these three classes by up to $10.4\%$ (highlighted in brown). This demonstrates that the normal dense CRFs model is effective in alleviating the ``broken-down'' error to some extent. However, the normal dense CRFs model harms the parsing results of small-scale objects. This can be observed from  $\text{IoU}_\text{class}$ of \emph{pole}, \emph{traffic light} and \emph{traffic sign} (highlighted in orange) which significantly drop \textit{w.r.t.}\  results provided by its baseline (baseline FCN + PEN fovea region). This is due to over-smoothness artifacts of the normal dense CRFs as it is unaware of the scale variance within the image.

In contrast, the perspective-aware CRFs model significantly boosts the $\text{IoU}_\text{class}$ of \emph{truck}, \emph{bus}, and \emph{train} by $15.7\%$, $6.0\%$, $12.3\%$ respectively (highlighted in red), without harming the results of small objects. Therefore, by incorporating perspective information, the perspective-aware CRFs model successfully reduces the ``broken-down'' errors without bringing over-smoothness, superior to the normal dense CRFs.
{The depth-aware CRFs model is superior to the normal dense CRFs one, but inferior to the perspective-aware CRFs one. This demonstrates that considering perspective geometry is useful but  depth prediction is not so discriminative as perspective information predicted by our proposed model, as discussed in Section \ref{subsec:pen}.}

Figure~\ref{fig:qualitative_crf} gives additional parsing examples from the perspective-aware CRFs model.  The trained Faster R-CNN model provides several object bounding boxes for the three urban scene images. PEN predicts perspective scores  on these objects, where a  brighter value indicates a higher probability of being near to the vanishing point (2nd row).
We can observe that before applying perspective-aware CRFs, large scale objects suffers from ``broken-down'' errors (3rd row). Perspective-aware CRFs  significantly reduces such errors in the peripheral region without over-smoothing small objects (\emph{e.g.}, \emph{pole}) (4th row).

\vspace{-4mm}
\paragraph{Comparison with State-of-the-art} 
We fine-tune the FoveaNet using both training and validation images. 
Then on the test set we compare its  performance with state-of-the-art published models which achieved best performance. Table~\ref{table:cityscapes_test} shows the results. 
Our FoveaNet  outperforms all the published state-of-the-arts.
FoveaNet performs especially well at instance-level (see iIoU results). Compared with the FCN model, FoveaNet brings significant improvement on $\text{iIoU}_\text{class}$ and $\text{iIoU}_\text{category}$, up to $5.2\%$. These two instance-level scores reflect the good parsing performance of FoveaNet on small scale objects. The improvement of $\text{IoU}_\text{class}$ and $\text{IoU}_\text{category}$ can be largely attributed to our perspective-aware CRFs, which can significantly reduce ``broken-down'' errors. Upon acceptance, we will release the code and model.

\begin{table}[t]
	\vspace{-2mm}
	\centering
	\footnotesize
	\caption{\small Performance comparison with baseline models on Cityscapes \emph{test} set.}
	\vspace{-3mm}
	\label{table:cityscapes_test}
	\setlength{\tabcolsep}{2pt}
	
	\begin{tabular}{l|ll|ll}
		\multirow{2}{*}{\raisebox{-0.1cm}{Methods}} & \multicolumn{2}{c|}{Class} & \multicolumn{2}{l}{Category} \\ \cline{2-5} 
		& \multicolumn{1}{l}{IoU} & \multicolumn{1}{l|}{iIoU} & \multicolumn{1}{l}{IoU} & iIoU \\
		\hline
		Dilation10 \cite{YuKoltun2016}    &   67.1    &     42.0      &      86.5        &  71.1   \\ 
		NVSegNet \cite{badrinarayanan2015segnet} &   67.4       &  41.4      &    87.2      & 68.1         \\ 
		DeepLabv2-(Resnet-101) \cite{liang2015semantic}  &    70.4     &  42.6    &  86.4     &   67.7     \\ 
		AdelaideContext \cite{lin2015efficient} & 71.6    & 51.7 &  87.3           &   74.1         \\ 
		LRR-4x \cite{ghiasi2016laplacian} &  71.8    &  47.9 &  88.3          &      74.1         \\ 
		\hline
		Baseline FCN &  71.3    &  47.2  &  87.8   & 72.9 \\
		FoveaNet (ours)   &  \bf{74.1}    &  \bf{52.4} &  \bf{89.3}       &     \bf{77.6}      
		
	\end{tabular}
	\vspace{-8mm}
\end{table}


\subsection{Results on CamVid}
\vspace{-2mm}
Cambridge-driving Labeled Video Database (CamVid)~\cite{BrostowFC:PRL2008} consists of over $10$min of high quality videos. There are pixel-level annotations of 701 frames with resolution $960\times 720$. Each pixel is labeled with one of the 32 candidate classes. Perspective geometry can also be observed on these frames. Following  previous works~\cite{badrinarayanan2015segnet,Ladicky:2010:MCO:1888089.1888122}, we use 11 classes for evaluation and report the per-pixel and average per-pixel accuracy.
To implement FoveaNet, we reuse PEN and Faster-RCNN trained on Cityscapes urban scene images. The two-branch FCN model (coarse and fovea branch) are initialized from ResNet-101 and fine-tuned on CamVid training and validation sets. The performance of FoveaNet on the test set and the comparison with state-of-the-arts are shown in Table~\ref{table:camvid_test}. FoveaNet outperforms the best baseline method on this dataset by $1.7\%$ and $3.7\%$ in global accuracy and average accuracy respectively. Due to limited space, we defer qualitative results on CamVid to Supplementary Material.

\begin{table}[t]
\vspace{-5mm}
\centering
\footnotesize
\caption{\small Performance comparison with baseline models on CamVid test set.}
\vspace{-3mm}
\label{table:camvid_test}
\setlength{\tabcolsep}{2pt}
\begin{tabular}{l|c|c}

      &   Global Accuracy    &  Average Accuracy      \\ \hline
  Zhang et al.\cite{zhang2012efficient}   &    82.1     &   55.4    \\ 
  Bulo et al.\cite{rota2014neural} &  82.1   &     56.1     \\ 
  Shuai et al.\cite{shuai2016dag} &  91.6    &  78.1         \\ 
  FoveaNet & \bf{93.3}  & \bf{81.8}
\end{tabular}
\vspace{-6mm}
\end{table}

\vspace{-2mm}
\section{Conclusion}
\label{sec:conclusion}
\vspace{-2mm}
We proposed a new urban scene parsing model FoveaNet by considering the ubiquitous scale heterogeneity when parsing scene images, which can provide state-of-the-art performance as validated on the Cityscapes and CamVid datasets. FoveaNet exploits the perspective geometry information through two novel components, perspective-aware parsing net and perspective-aware CRFs model, which work jointly and successfully to solve the common scale issues, including parsing errors on small distant objects, ``broken-down'' errors on large objects and over-smoothing artifacts. 

\vspace{-2mm}
\section{Acknowledgments}
\label{sec:acknowledgments}
This work was supported by the National Natural Science Foundation of China under Grant No. 61471214 and the National Basic Research Program of China (973 program) under Grant No.2013CB329403. The work of Jiashi Feng was partially supported by NUS startup R-263-000-C08-133, MOE Tier-I R-263-000-C21-112 and IDS R-263-000-C67-646.

{\small
\bibliographystyle{ieee}
\bibliography{egbib}
}

\end{document}